\documentclass[letterpaper, 10 pt, conference]{ieeeconf}  % Comment this line out if you need a4paper

\IEEEoverridecommandlockouts
\usepackage[linesnumbered,lined,boxed,ruled]{algorithm2e}
\usepackage{subcaption}
\usepackage{adjustbox}
\usepackage{booktabs}
\usepackage{amsmath}                               
\usepackage{amssymb}  
\usepackage{color, soul}
\usepackage{multirow}

\usepackage[bookmarks=true]{hyperref}

\newcommand{\tm}[1]{\text{#1}}

\DeclareMathOperator*{\argmin}{arg\,min}

\title{\LARGE \bf
Fast maneuver recovery from aerial observation:\\ trajectory clustering and outliers rejection
}

\author{Nelson de Moura$^{1}$, Augustin Gervreau-Mercier$^{1, 2}$, Fernando Garrido$^{3}$ and Fawzi Nashashibi$^{1}$% <-this % stops a space
	\thanks{$^{1}$Nelson de Moura, Fawzi Nashashibi and Augustin Gervreau-Mercier are with INRIA, 75012 Paris, France,
		{\tt\small \{nelson.demoura;~fawzi.nashashibi\}@inria.fr}}%
	\thanks{$^{2}$Augustin Gervreau-Mercier is also with Université Paris Dauphine-PSL, 75775 Paris, France,
		{\tt\small  augustin.gervreau@dauphine.eu}}%
	\thanks{$^{3}$Fernando Garrido is with Valeo DSW team, 94000 Créteil, France,
		{\tt\small fernando.guarrido@valeo.com}}%
	\thanks{This research has been funded by the plan "France Relance", grant agreement number ANR-21-PRRD-0005-01}		
}

\begin{document}

\maketitle
\thispagestyle{empty}
\pagestyle{empty}

%%%%%%%%%%%%%%%%%%%%%%%%%%%%%%%%%%%%%%%%%%%%%%%%%%%%%%%%%%%%%%%%%%%%%%%%%%%%%%%%
\begin{abstract}

The implementation of road user models that realistically reproduce a credible behavior in a multi-agent simulation is still an open problem. A data-driven approach is proposed here to infer behaviors that may exist in real situation to obtain different types of trajectories from a large set of observations. The data, and its classification, could then be used to train models capable to extrapolate such behavior. Cars and two different types of Vulnerable Road Users (VRU) will be considered by the trajectory clustering methods proposed: pedestrians and cyclists. The results reported here evaluate methods to extract well-defined trajectory classes from raw data without the use of map information while also separating "eccentric" or incomplete trajectories from the ones that are complete and representative in any scenario. Two environments will serve as test for the methods develop, three different intersections and one roundabout. The resulting clusters of trajectories can then be used for prediction or learning tasks or discarded if it is composed by outliers. 

\end{abstract}

%%%%%%%%%%%%%%%%%%%%%%%%%%%%%%%%%%%%%%%%%%%%%%%%%%%%%%%%%%%%%%%%%%%%%%%%%%%%%%%%
\section{Introduction}

Simulation is a indispensable tool to prove the efficacy and viability of any framework or system capable to drive an Automated Vehicle (AV) before integration with a prototype. In most cases these simulations reproduce the behaviors of other road users based on real recordings, which in this case the behavior of all road users is fixed ahead of time, or it relies on hybrid approaches, mixing real information with some \textit{a priori} hypothesis, modeling and/or knowledge about the agents being represented. The main goal of this work is to produce classifications of trajectories to feed these hybrid methods with reliable and diverse set of trajectories, for different situations and different road users while being fast to execute and reliable enough to sift through outliers at input.  

Trajectory clustering has been a long research topic in the AV area. Many articles deal specially with the analysis of vehicle trajectory as a way to retrieve the possible trajectories in an environment, to study the traffic flow intersections \cite{choong2017modeling}, to discover possible longitudinal behaviors of vehicles \cite{demoura2023extraction}, to execute some learning task \cite{wang2021vehicle} or even to examine the scenarios that might happen during driving \cite{ries2021trajectory}. Thus, the goal of this paper is to propose a fast and robust way to recover sets of trajectories for vehicles and vulnerable road users (VRU), delivering sets of trajectory samples as input to all the aforementioned tasks in a simple manner. 

Approaches based on clustering with dynamic time warping (DTW) are the norm in the literature. In \cite{ries2021trajectory} three different threshold comparisons were made using the DTW distance metric to establish a similarity relationship between scenarios involving multiple road users. K-means and fuzzy c-means were used for \cite{choong2017modeling, choong2018modeling} with longest common subsequence (LCSS) to cluster trajectories in intersections so to derive insights about the traffic flow in multiple lane cross-intersections. Changing from the urban to aerial traffic, \cite{liu2021clustering} proposed a method to combine k-means with outlier removal based on information theory by the minimization of the holoentropy and achieving good results clustering flight trajectories. With a different motivation, \cite{demoura2023extraction} implemented a fast k-means clustering method only for vehicles and bypassing the outliers problem. And on a totally different scale \cite{wang2021vehicle} applied the same idea of trajectory clustering but on a city scale, learning an embedding to simplify the trajectory representation and then clustering the projected data to find vehicles with similar behavior.

\begin{figure}[t]
	\centering
	\begin{subfigure}[b]{0.49\columnwidth}
		\centering
		\adjustbox{scale=1.2, trim=20mm 16mm 19mm 7.5mm, clip}{
			\includegraphics[width=1.8\textwidth]{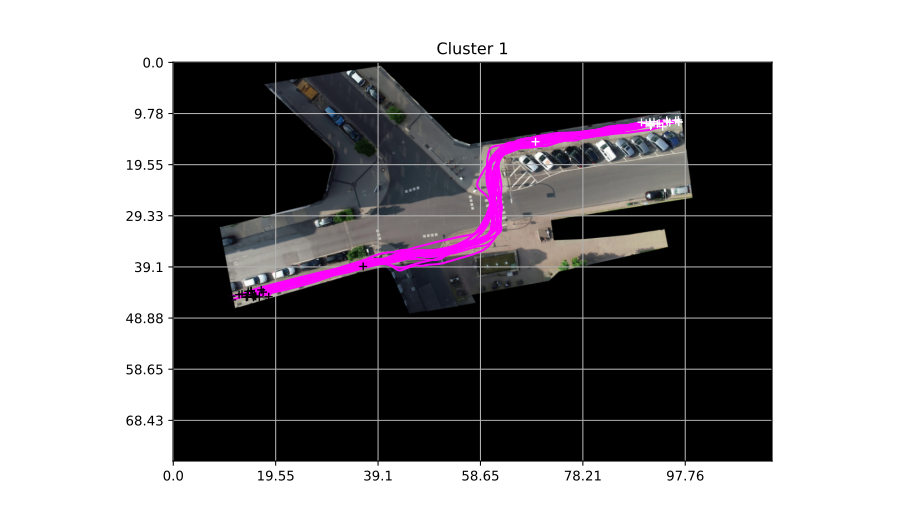} 
		}
		\caption{Pedestrian on sce. 2 of inD}
		\label{fig:10a}
	\end{subfigure}

	\begin{subfigure}[b]{0.49\columnwidth}
		\centering
		\adjustbox{scale=1, trim=25mm 20mm 25mm 5mm, clip}{
			\includegraphics[width=2\textwidth]{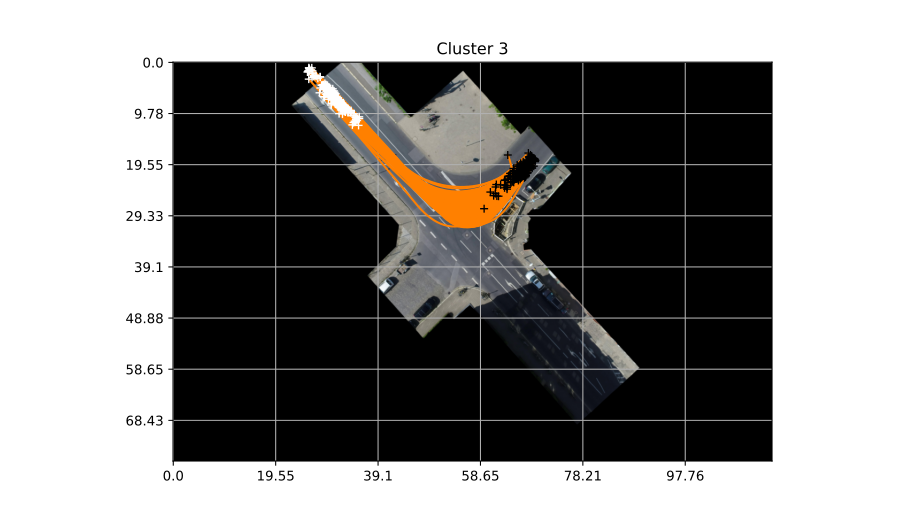} 
		}
		\caption{Vehicles on sce. 1 of inD}
		\label{fig:10c}
	\end{subfigure}
	\begin{subfigure}[b]{0.49\columnwidth}
		\centering
		\adjustbox{scale=1, trim=22.5mm 8mm 30mm 15mm, clip}{
			\includegraphics[width=2\textwidth]{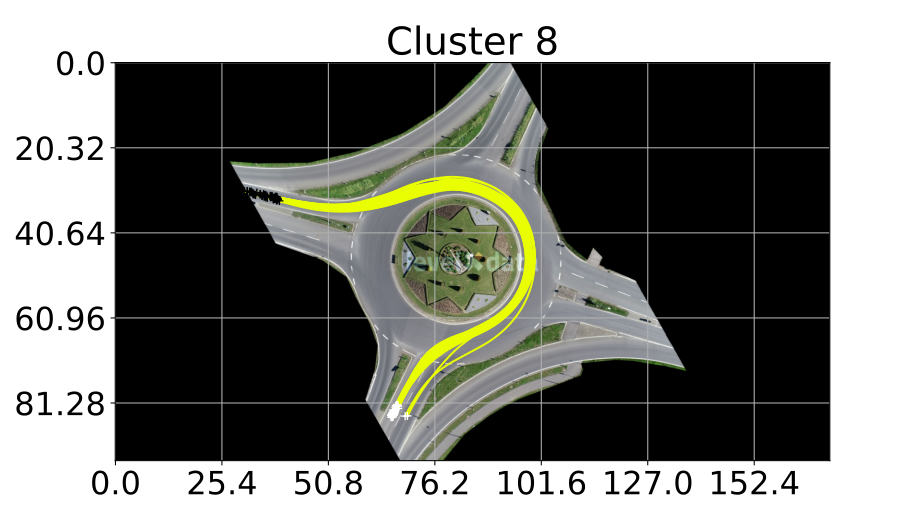} 
		}
		\caption{Vehicles on sce. 2 of rounD}
		\label{fig:10d}
	\end{subfigure}
	\caption{Trajectory cluster for different road users}
	\label{fig:10}
\end{figure}

The contribution of this paper is two-fold: First, to propose a fast clustering method for maneuver retrieval from real observations that do not need any map information and that is compatible with vehicles and VRUs. Second, to adapt this model to deal with trajectories that can be considered as outliers, separating these "eccentric"\footnote{Definition on section \ref{sec:2}} and/or erroneous instances without disturbing the clustering process. A preview of the results obtained can be seen in Figure \ref{fig:10}. Given the simplicity of the approach, the method presented here can be a valuable addition to the plethora that already exists.

All data used originate from a microscopic observation of road environments by a drone \cite{bock2020, krajewski2020round}, which retain a good amount of information about the behaviors of each road users, in comparison with \cite{barmpounakis2020pneuma} that make observations on a larger scope and well structured environment (4-lane road with signals), and of \cite{zhan2019interaction} that focus more on the interaction size with short observation periods (4s) and $10\tm{Hz}$ of acquisition frequency (in comparison with $25\tm{Hz}$ of the data used here).  

\section{Separating trajectories of interest}
\label{sec:2}

\begin{figure}[h]
	\centering
	\begin{subfigure}[b]{0.49\linewidth}
		\centering
		\adjustbox{scale=2.5, trim=20mm 7.5mm 15mm 10mm, clip}{
			\includegraphics[width=\textwidth]{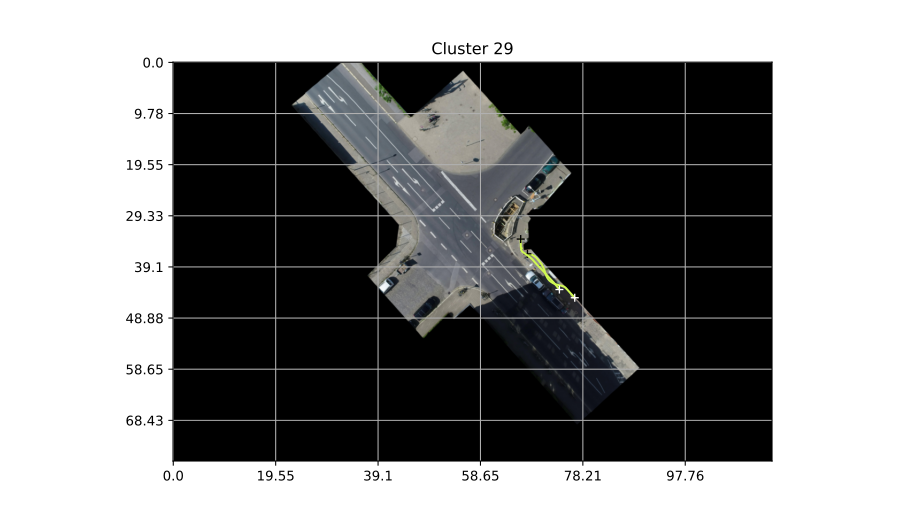} 
		}\caption{Eccentric}
		\label{fig:1a}
	\end{subfigure}
	\begin{subfigure}[b]{0.49\linewidth}
		\centering
		\adjustbox{scale=2, trim=17.5mm 10mm 15mm 3mm, clip}{
			\includegraphics[width=\textwidth]{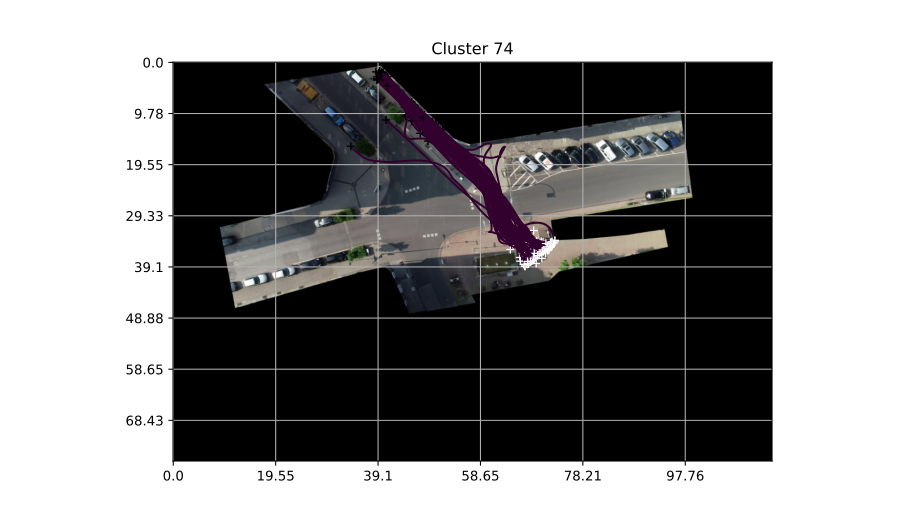} 
		}
		\caption{Target}
		\label{fig:1b}
	\end{subfigure}
	\caption{One eccentric trajectory and a target cluster (background image from \cite{bock2020})}
	\label{fig:1}
\end{figure}

Differently from the vehicle trajectories studied in \cite{demoura2023extraction}, VRU trajectories are less constraint by its environment and are also more prone to acquisition error as well (shadows, changes in direction, multiple users close by). Also, when a scenario for observation is defined some of the less intuitive trajectories become superfluous, considering the interest of keeping only those which represent behaviors that can be transposed in other scenarios. Take, for example, the trajectories displayed in Figure \ref{fig:1} (white cross represents the beginning of the trajectory and black cross the end): these three different sets may represent a real-life situation, like getting out of a store and entering in a car but they are not of interest since they are scenario-specific. These types of trajectories will be qualified as \textit{eccentric} from now on. 

The main focus is to sift through entire datasets and isolate the eccentric trajectories (like the ones in Figure \ref{fig:1a}) and erroneous ones (like vehicles starting their trajectory in the middle of an intersection) in specific clusters and trajectories of interest Figure \ref{fig:1b} in their own clusters. The final result can then be visually inspected to discard some and retain others. Both figures were produced using the data available in the InD Dataset \cite{bock2020}.

Given the difficulties that methods of the similar inspiration of k-means have with outliers, other clustering algorithms were considered to deal with pedestrian and cyclist trajectories. All of them using a pre-calculated dissimilarity matrix (Equation \eqref{eq:17}) where each element is the result of the Dynamic Time Warping (DTW, subsection \ref{subsec:2a}) distance measure of two trajectories.

\begin{equation}
	\label{eq:17}
	\mathcal{D}_{DTW} = \begin{bmatrix}
		d_{0,0}^{DTW} = 0	& \cdots	&	d_{0, n}^{DTW}\\
		d_{1,0}^{DTW}		& \cdots	&	d_{1, n}^{DTW}\\
		\vdots				& \ddots	&	\vdots\\
		d_{n,0}^{DTW}		& \cdots	&	d_{n, n}^{DTW} = 0
	\end{bmatrix}
\end{equation} 

\subsection{Dynamic time warp (DTW)}
\label{subsec:2a}

DTW was first introduced in the speech processing domain as a way to compare two time series that have different phases. Even though the trajectories studied here were sampled at the same frequency, they may have different lengths and although corresponding to the same maneuver in an intersection. Consider two discrete time series, represented by \eqref{eq:3} and \eqref{eq:4}, with different sizes $n$ and $m$, where \begin{math}
	K = \left\{k_0, k_1, \ldots, k_n, \ldots, k_m, \ldots\right\}
\end{math} represents the sampled periods:

\begin{eqnarray}
	\label{eq:3}
	R[K] = r[k_0], r[k_1], \ldots, r[k_n]\\
	\label{eq:4}
	S[K] = s[k_0], s[k_1], \ldots, s[k_m]
\end{eqnarray}

The goal of the DTW is to calculate the optimal sequence of pairs of point indexes, one from each time series. This is done by minimizing the euclidean distance between the points indicated by the index pair, from $(r[k_0], s[k_0])$ to $(r[k_n], s[k_m])$, using a certain set of increments to walk from the former to the latter. In the standard implementation (equation \ref{eq:2}) three steps are tested: +1 on the index 1, +1 on the index 2 or +1 in both. Equation \eqref{eq:1} defines the DTW from R and S as the calculated sum of distances, which are determined by the recursion in Equation \eqref{eq:2}, for $0 \leq i \leq k_n$ and $0 \leq j \leq k_m$. 

\begin{equation}
	\label{eq:1}
	DTW(R,S) = \gamma(k_n,k_m)
\end{equation}
\vspace{-7.5mm}
\begin{multline}
	\label{eq:2}
	\gamma(i,j) = d(r[k_i], s[k_j]) +\\ \min\left[\gamma(i-1,j), \gamma(i,j-1), \gamma(i-1,j-1)\right]
\end{multline}

There are multiple DTW variants, some changing the walk used in the recursion \eqref{eq:2} (constraint DTW \cite{muller2007information}) or adopting restriction on the elements to be considered by Equation \eqref{eq:2} (Sakoe-Chiba band \cite{sakoe1978dynamic}; Itakura parallelogram \cite{itakura1975minimum}). Usually, when the Euclidean metric is used, the centroid of a set of series can be calculated simply by summing all the elements and dividing by the number of series in the set. One can do the same with series of different lengths using the Dynamic Barycenter Averaging (DBA) \cite{petitjean2011global}, however the result of this algorithm usually is a non-differentiable array always with the same size of the biggest array in the set. When a time-series is necessary to represent the ensemble of a cluster, the medoid will be chosen, according to Equation \eqref{eq:5}, where $\mathcal{X}$ is the set being considered and $d$ in our case is the DTW distance. 

\begin{equation}
	\label{eq:5}
	x_{\tm{med}} = \argmin_{x\in\mathcal{X}}\sum_{i=0}^{N}d_{\tm{DTW}}(x, x_i)
\end{equation}

\subsection{Clustering methods}
\label{subsec:2b}

Three main methods were used to cluster the trajectories using the DTW distance: hierarchical clustering, partition around medoids (or k-medoids) and dissimilarity matrix clustering. 

\subsubsection{Hierarchical clustering}
\label{subsubsec:2b1}

The hierarchical clustering used was based on a agglomerative processes, i.e., it starts with each sample being a cluster and at each step it merges the two most similar clusters, to then continue this process until the desired number of clusters is achieved \cite{schutze2008introduction}. The metric used to measure the similarity of two clusters, thus to decide with clusters should be merged at a given iteration, was the average linkage, Equation \eqref{eq:6}:

\begin{equation}
	\label{eq:6}
	d_{\mathcal{C}_i, \mathcal{C}_j} = \frac{1}{N\cdot M}\sum_{x_i\in \mathcal{C}_i}^{N}\sum_{x_j\in \mathcal{C}_j}^{M}d(x_i, x_j)
\end{equation}

Where $\mathcal{C}_i$ and $\mathcal{C}_j$ are two clusters being evaluated and $N$ and $M$ are the number of elements inside each respective cluster. The distance measure used is the DTW (from \eqref{eq:17}).

\subsubsection{Partition around medoids (or k-medoids)}
\label{subsubsec:2b2}

It uses the same sequence of calculation - allocation of elements in cluster then center recalculation - from the k-means algorithm but using the medoid element as the cluster center (equation \ref{eq:5}), not a synthetic average of elements \cite{kaufman2009finding}. Such adaptation is common in cases where it is difficult to calculate the average of elements being clustered, as for example when these do not have the same length. 

\subsubsection{Dissimilarity matrix clustering}
\label{subsubsec:2b3}

The algorithm was proposed to accelerate the clustering of vehicle's trajectories in \cite{demoura2023extraction}, in comparison with the k-means algorithm. To simplify the k-means calculation, it is applied at each row of the dissimilarity matrix for the entire dataset to look for the smallest cluster center of all rows (which will be the one that has the smallest sum of distances to the element represented by the row). Then, all the elements assigned to this minimal cluster are removed from the matrix and the process is executed again, until the desired number of clusters is achieved. If there are any left, then they are assigned to the cluster in which it has the smallest distance to its medoid. 

\section{Methods evaluated}
\label{sec:3}

Independently from the clustering method used from the proposed in the previous section, some errors in classification might still appear, even if a higher number of clusters is used. To cluster using the DTW distance considers only the shape of the trajectory, which might mix trajectories that have small but important differences at its origin or terminus but that share an important part of its path. Hence, to correct this errors the initial and final points shall be used in a separate clustering procedure split the elements based on these points. Then, it will be necessary to check if any of the just-obtained sub-clusters should be fused back together, if they really are on the same maneuver, or even if they should be merged with other sub-clusters, to determine the final result. Algorithm \ref{alg:1} shows how the entire clustering process with this post-processing operation works. The interval $[nk_{\tm{min}}, nk_{\tm{max}}]$ refers to the minimal and maximal number of clusters to be evaluated. 

\begin{algorithm}
	\caption{Proposed clustering procedure}
	\label{alg:1}
	\For{$n_k \in [nk_{\tm{min}}, nk_{\tm{max}}]$  }{
		clusters, centers = \textbf{agglo\_clustering}(data)\\
		div\_cl = \textbf{split\_clusters}(clusters)\\
		final\_cl, final\_centers = \textbf{merge\_clusters}(div\_cl)
	}
\end{algorithm}

\subsection{Reorganization using initial and final points}
\label{subsec:3a}

Two different approaches were taken to evaluate the best solution to further divide the clusters according to their initial and final points:

\begin{itemize}
	\item Cluster both initial and final points on the same array, establishing automatically the sub-cluster groupings
	\item Cluster the initial point, then the final point and list the grouping created comparing both results.
\end{itemize} 

Given that a search based on a number of clusters is already being executed, the mean-shift algorithm was used to execute these two options of post-processing, avoiding a nested search. After the merge process will happen, to fuse clusters with fairly similar characteristics, only differing from a few meters from each other while retaining its significant part, for example a left turn, a street crossing, etc. 

Evaluating if two sub-clusters should be merged is done using the medoid of each cluster, together with the spread of elements in each sub-cluster, Equation \eqref{eq:9}. The variable $m_i$ represents the medoid of the cluster $\mathcal{C}_i$ and $N_i$ its number of trajectories. To merge one sub-cluster with another it is necessary that the distance between both medoids be smaller than the sum of spreads for the respective clusters (line \ref{alg2:spread} of algorithm \ref{alg:2}). To discard small differences, the medoid of one cluster is projected into the other (line \ref{alg2:proj} of algorithm \ref{alg:2}), so that the similarity disregard any errors in tracking or even small differences in the start or terminus of the trajectory. If this condition is true, there is another to be fulfilled: the calculated projection should be equal or higher than a certain percentage of the original trajectory (which is defined in Table \ref{tab:11} for all cases examined here).

\begin{equation}
	\label{eq:9}
	\tm{spr}_{\mathcal{C}_i} = \frac{1}{N_i}\sum_{x_i\in \mathcal{C}_i}d_{\tm{DTW}}(m_i, x_i)
\end{equation}

\begin{algorithm}
	\SetAlgoNoEnd
	\caption{Merging sub-clusters}
	\label{alg:2}
	\For{Cluster i in $\mathcal{C}$}{
		\For{Cluster j $\neq$ i in $\mathcal{C}$}{
			$\tm{proj}_{i\rightarrow j}$ = \textbf{Projection}($\tm{medoid}_i$, $\tm{medoid}_j$)\label{alg2:proj}\;
			$d_\tm{proj}^{i, j}$ = DTW($\tm{med}_j$, $\tm{proj}_{i\rightarrow j}$)\;
			\If{$d_\tm{proj}^{i, j} \leq \tm{spr}_i + \tm{spr}_j$}{\label{alg2:spread}
				\lIf{$\tm{\textbf{trace}}_{i} \leq c_{min_fac}\cdot \tm{\textbf{trace}}_{j}$}{$\mathcal{C}_j \leftarrow \mathcal{C}_i$}
				\Else{
					\textbf{continue}
				}
			}
		}
	}
	
\end{algorithm}

Equations \eqref{eq:7} and \eqref{eq:8} show how one trajectory is projected onto another. For two generic trajectories $\tm{\textbf{tr}}_a = (p_{a0}, p_{a1}, \cdots, p_{a_n})$ and $\tm{\textbf{tr}}_b = (p_{b0}, p_{b1}, \cdots, p_{b_m})$, two loops are executed: one for the initial point of ${\tm{\textbf{tr}}}_a$ and another its final point. Inside these loops the index $j$ in increased from 0 (or decreased from $N_b$ for the terminus) to find the interval of points where $p_{a_0}$ (or $p_{a_n}$) is perpendicularly projected. The cutting point is obtained when $\lambda_{b_j} \in [0,1]$, meaning that the projection of $p_{a0}$ (or $p_{an}$), is between points $j$ and $j+1$ from $\tm{\textbf{tr}}_b$. If no cutting point is detected then all the trajectory is used in the comparison. 

\begin{gather}
	\label{eq:7}
	v_{p_{a_0} \rightarrow p_{b_j}} = p_{b_j} - p_{a_0}\\
	\label{eq:8}
	\lambda_{b_j} = \frac{\left[v_{p_{a_0} \rightarrow p_{b_j}}\right] \cdot \hat{v}_{p_{b_{j, j+1}}}}{\|\hat{v}_{p_{b_{j, j+1}}}\|}
\end{gather} 

\subsection{Evaluating cluster partition quality}
\label{subsec:3c}

Finally, it is necessary to evaluate the clusters according to the similarity of elements inside each cluster and in other clusters as well. Three metrics will be used for it: the Davies-Bouldin index the Silhouette score and the spread on cluster, proposed here. The DB index will be slightly modified to allow a better representation of the distribution quality for a certain number of clusters, while the latter will be used as it was defined in \cite{rousseeuw1987silh}. 

\subsubsection{Davies-Bouldin index (\textbf{DB})}
\label{subsubsec:3a1}

The Davies-Bouldin index (DB) is originally defined as the average of the maximal value of $R_{ij}$, as if defined in Equations \eqref{eq:10} and \eqref{eq:11}. Equation \eqref{eq:9} is used to calculate the spread $s_i$. Since it is the maximal value of $R_{ij}$ that is used to calculate the final score, it has creates a dependency to the number of clusters, i.e. the decrease on the score is connected to a higher number of cluster and not necessarily a better distribution. 

\begin{gather}
	\label{eq:10}
	R_{ij} = \frac{\left(s_i+s_j\right)}{d_{ij}}\\
	\label{eq:11}
	\tm{DB}_{n_c} = \frac{1}{n_c}\sum_{i=0}^{n_c}\left[\max_{j=1,\ldots,n_c, i\neq j}R_{ij}\right]
\end{gather}

Thus, a small modification was done, to use the average of $R_{ij}$, not its maximal value, as it can be seen in Equation \eqref{eq:12}. This offers less bias to the number of clusters, considering always all the distribution of elements being evaluated. The $n_c - 1$ discounts the distance of the medoid to itself, which is zero.

\begin{gather}
	\label{eq:12}
	\tm{DB}_{n_c} = \frac{1}{n_c}\frac{1}{n_c - 1}\cdot \sum_{i=0}^{n_c}\sum_{j=0}^{n_c}R_{ij}
\end{gather}

\subsubsection{Silhouette score (\textbf{Slh.})}
\label{subsubsec:3a2}

Another metric to evaluate the clustering quality is the silhouette score, proposed in \cite{rousseeuw1987silh}. Differently from the DB score, it is calculated for each element being clustered, with $a(i)$ begin the average dissimilarity (DTW distance in this case) of element $i$ to all other elements in its cluster.  The other value necessary to calculate the silhouette is the minimum average dissimilarity between the element in question to the other clusters, Equation \eqref{eq:15}. 

\begin{gather}
	\label{eq:14}
	s_{\mathcal{C}_i}(i) = \frac{b(i) - a(i)}{\max\left(a(i), b(i)\right)}\\
	\label{eq:15}
	b(i) = \min_{\mathcal{C}_j \neq \mathcal{C}_{i}}d_{\tm{DTW}}\left(b(i), \mathcal{C}_j\right)
\end{gather}

With the $s(i)$ for each trajectory, the silhouette score for the clustering is the average score for all elements. This score is contained in $[-1, 1]$, with a score close to 1 being excellent (distances inside cluster are much smaller than distance between clusters). As the DB score it compares a infra-cluster spread measure with inter-cluster distances, but in a individual level. As it will be seen, in some situations where many different clusters co-exist close to each other (subsection \ref{subsec:4.2}), it will not be a representative measure of cluster quality.    

\subsubsection{Spread on cluster (\textbf{Spr.})}
\label{subsubsec:3a3}

This measure is somewhat similar to Equation \eqref{eq:7} but was changed to capture the biggest difference between two members of the same cluster. In Equation \eqref{eq:16} the average of all the spreads divided by the number of members in the respective cluster define the metric.

\begin{equation}
	\label{eq:16}
	\theta_{\mathcal{C}} = \frac{1}{\|\mathcal{C}\|}\cdot \sum_{i=0}^{\|\mathcal{C}\|} \frac{\max_{j, k\in \mathcal{C}_i}\left[d_{\tm{DTW}}(x_j, x_k)\right]}{\|\mathcal{C}_i\|}
\end{equation}

This metric will be specially important for the pedestrian case, where the silhouette score is not representative given the close proximity of multiple clusters.

\section{Results}
\label{sec:4}

\subsection{Methodology}
\label{subsec:4.1}

Two different sources of data will be used to test the algorithms proposed here: the inD dataset \cite{bock2020} and the roundD dataset \cite{krajewski2020round}. Both are obtained using a unnamed aerial vehicle (UAV), the former containing four different intersections and the latter three roundabouts, all in Germany. The information about the number of trajectories per scenario and per road user can be found in table \ref{tab:1}; the number of files refers to the data-batch file division for each scenario: in the inD case only the first three intersections were used for the VRU (for the vehicle case all intersection were tested) and for the roundD there are two files with two different roundabouts that are not used (only a few observations); the other scenarios are obtained from the observation of a third one (files 02 to 23). This data was split into three scenarios, given the number of trajectories (the clustering results of the three scenarios could be merged using the algorithm proposed in \cite{demoura2023extraction}). For the clustering execution all three road user trajectories from inD were used (in blue at Table \ref{tab:1}), while for the rounD only the vehicles' trajectories could be used, due to the low number of observations for pedestrians and cyclists (in red at Table \ref{tab:1}).

\begin{table}[h]
	\caption{Number of trajectories per scenario per road user} 
	\label{tab:1}
	\centering
	%\scriptsize
	\begin{tabular}{c|ccc}
		\toprule
		\textbf{Scenario} 			& \textbf{Cars} & \textbf{Pedestrians} 	& \textbf{Cyclists} \\
		\midrule
									& \multicolumn{3}{c}{inD Dataset}\\
		\midrule
		Sce. 0 - Files 00 to 06 	& \color{blue}1826 	& \color{blue}144	& \color{blue}83 	\\
		Sce. 1 - Files 07 to 17 	& \color{blue}2337 	& \color{blue}770	& \color{blue}420	\\
		Sce. 2 - Files 18 to 29 	& \color{blue}2133	& \color{blue}2005	& \color{blue}1669	\\
		Sce. 3 - Files 30 to 32 	& \color{blue}1098 	& \color{red}38		& \color{red}39		\\
		\midrule
									& \multicolumn{3}{c}{rounD Dataset}\\
		\midrule
		Sce. 0 - Files 02 to 08		& \color{blue}3819 	& \color{red}4		& \color{red}20	\\
		Sce. 1 - Files 09 to 15		& \color{blue}2984 	& \color{red}5		& \color{red}21 \\
		Sce. 2 - Files 16 to 23		& \color{blue}3983 	& \color{red}2		& \color{red}32 \\		
		\bottomrule
	\end{tabular} 
\end{table}

Table \ref{tab:11} shows the parameters used for the tests that will be presented next. The number of cluster established the interval clusters tested by the methods, while the \textit{Min. trace} refers to the minimal percentage that the projected medoid must have to be merged with another cluster (subsection \ref{subsec:3a}, algorithm \ref{alg:2}).

\begin{table}[h]
	\caption{Parameters used in clustering} 
	\label{tab:11}
	\centering
	\scriptsize
	\adjustbox{max width=\columnwidth}{
		\begin{tabular}{c|ccc}
			\toprule
			\textbf{User} 				& \textbf{Scenario} & \textbf{Nb. clusters} 	& \textbf{Min. trace} 	\\
			\midrule
			\multirow{3}{*}{\textbf{Pedestrian}}	& inD sce. 0		& [5, 32]	& 0.6 \\
													& inD sce. 1		& [15, 45]	& 0.6 \\
													& inD sce. 2		& [70, 110]	& 0.7 \\
			\midrule
			\multirow{3}{*}{\textbf{Cyclist}}	& inD sce. 0		& [5, 20]		& 0.6 \\
												& inD sce. 1		& [15, 45]		& 0.6 \\
												& inD sce. 2		& [20, 60]		& 0.6 \\
			\midrule
			\multirow{5}{*}{\textbf{Car}}		& inD sce. 0		& [5, 15]		& 0.67 \\
												& inD sce. 1		& [5, 20]		& 0.6 \\
												& inD sce. 2		& [5, 28]		& 0.4 \\
												& inD sce. 3		& [5, 22]		& 0.45 \\
												& rounD sce. 0-2	& [10, 25]		& 0.6 \\
			\bottomrule
		\end{tabular}
	} 
\end{table}

The choice for this specific datasets was motivated because both of them are captured by drone, not by an automated vehicle in the environment, which could modify the behaviors observed and also because the metadata present in the dataset allows the trajectories to be plotted in a realistic background image. But the method presented here could be used in any other ensemble of trajectories. Also, all the data is used as-is, no trajectory in the dataset is discarded beforehand only a normalization is done on each dimension of the trajectory before calculating the dissimilarity matrix (Equation \ref{eq:17}). All the methods were implemented in Python.

\subsection{Pedestrians}
\label{subsec:4.2}

There are three intersection scenarios for pedestrians, all in the inD dataset. The most important problem with pedestrians is the high variability of maneuvers, because it have an almost constraint-free environment to evolve and also due to detection and tracking errors during the dataset acquisition and post-processing. It is a real challenge for the clustering process to treat all these problems and produce a compact cluster set. The results for scenarios 0 and 2 can be seen in table \ref{tab:2} and table \ref{tab:3}. 

The abbreviation \textit{Agglo} refers to the pure agglomerative clustering, A2MS to the agglomerative followed by two mean-shits, on the initial and final points separately and A1MS to one mean-shift on the initial and final points on the same array. For all tables, \ref{tab:2} through \ref{tab:10}, the time indicated is the average clustering time per cluster calculated. In the column best $n_k$ the first value is the nominal value of the number of clusters used, and in parenthesis it is the final number of clusters, and in the last column the percentage of the original trajectories present in the final classification (in parenthesis the number of rejected trajectories).

Both scenarios are two extremes for the clustering process: one has few trajectories and not many options for a pedestrian to evolve in the environment and the other has much more of both. It is exactly of possible destinations and the possibility to use the same space in both directions that make the silhouette measure to fail when choosing the best method. Since the number of optimal clusters for each method is different, the spread on cluster, defined in \ref{subsubsec:3a3} is the best choice of criteria to select the best method overall and in the current case it indicates that the agglomerative clustering with two mean-shift applications (A2MS) is the best option for the scenario 0 and 2 (for the 1 as well, for space limitations it will not be shown here). 

\begin{table}[h]
	\caption{Results for scenario 0 of inD dataset} 
	\label{tab:2}
	\centering
	\scriptsize
	\adjustbox{max width=\columnwidth}{
		\begin{tabular}{c|cccccc}
			\toprule
			\textbf{Method} & \textbf{Avg. time} & \textbf{DB}	& \textbf{Slh.} & \textbf{Spr.} & \textbf{Best $n_k$} & \textbf{Nb trajs.}\\
			\midrule
			Agglo. 		& 0.001 s	& 0.3164			& \textbf{0.7012}	& 1.7561 			&	32 (22) 	& 93.06\% (10)\\
			A2MS 		& 13.38 s	& \textbf{0.2546}	& 0.5753		 	& \textbf{1.4250}	&	23 (19)		& 79.86\% (29)\\
			A1MS 		& 10.36 s	& 0.2837			& 0.5823			& 1.5823			&	19 (18) 	& 79.86\% (29)\\
			PAM 		& 16.07 s 	& 0.5407			& 0.5810			& 3.2508			&	26 (24)		& 98.61\% (2)\\
			Dissi. 		& 16.68 s 	& 0.4820			& 0.4921			& 3.5506			&	32 (29)		& 97.92\% (3)\\
			\bottomrule
		\end{tabular} 
	}
\end{table}

For both scenarios the same observation can be made: the A2MS method has the lowest DB score and spread while the pure agglomerative method has the biggest silhouette score. This is exactly because the latter method some clusters that should be separated end up together while for the former one they usually can be separated by the second clustering. In both cases, clusters with a single element are discarded from the final distribution (and not accounted for during the calculation of the scores presented in each table). The exactly same observations can be made for the scenario 2\footnote{For lack of time to process the data, the results for the PAM and dissi. method are not available at submission time; it does not change the final conclusion. At the final submission the values will be updated.}. 

\begin{table}[h]
	\caption{Results for scenario 2 of inD dataset} 
	\label{tab:3}
	\centering
	\scriptsize
	\adjustbox{max width=\columnwidth}{
		\begin{tabular}{c|cccccc}
			\toprule
			\textbf{Method} & \textbf{Time} & \textbf{DB}	& \textbf{Slh.} & \textbf{Spr.} & \textbf{Best $n_k$} & \textbf{Nb trajs.}\\
			\midrule
			Agglo. 	& 0.04 s	& 0.4751			& \textbf{0.5330}	& 1.7863 			&	108 (77) 	& 98.45\% (31)\\
			A2MS 	& 57.75 s	& \textbf{0.3500}	& 0.4693		 	& \textbf{1.3265}	&	81 (77)		& 94.96\% (101)\\
			A1MS	& 34.90 s	& 0.4257			& 0.4610			& 1.6738			&	72 (77) 	& 96.71\% (66)\\
			PAM 	& - 	& -			& -			& -			& -			& \\
			Dissi. 	& - 	& -			& -			& -			& -			& \\
			\bottomrule
		\end{tabular}
	} 
\end{table}

\begin{figure}
	\centering
	\begin{subfigure}[b]{0.49\linewidth}
		\centering
		\adjustbox{scale=1.4, trim=35mm 15mm 22.5mm 20mm, clip}{
			\includegraphics[width=2\textwidth]{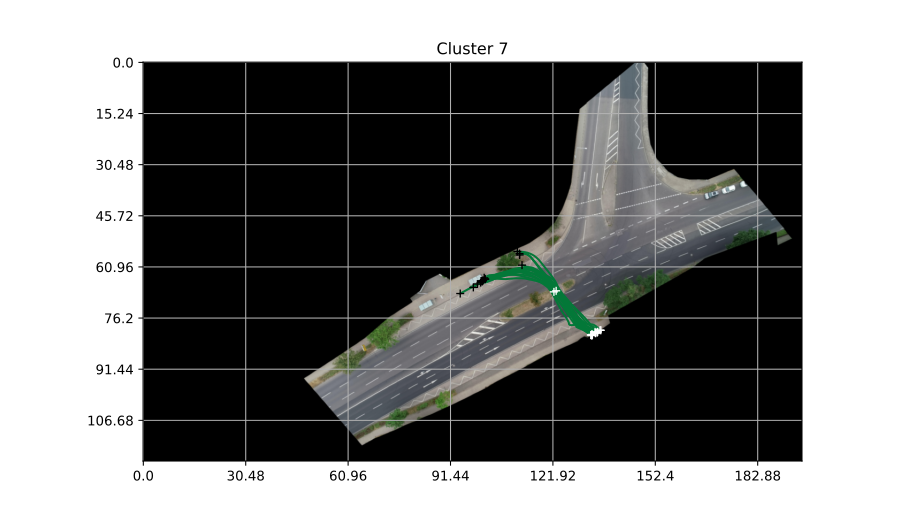} 
		}
		\caption{Agglo method clust. 2}
		\label{fig:4a}
	\end{subfigure}
	\begin{subfigure}[b]{0.49\linewidth}
		\centering
		\adjustbox{scale=1.2, trim=35mm 15mm 22.5mm 20mm, clip}{
			\includegraphics[width=2\textwidth]{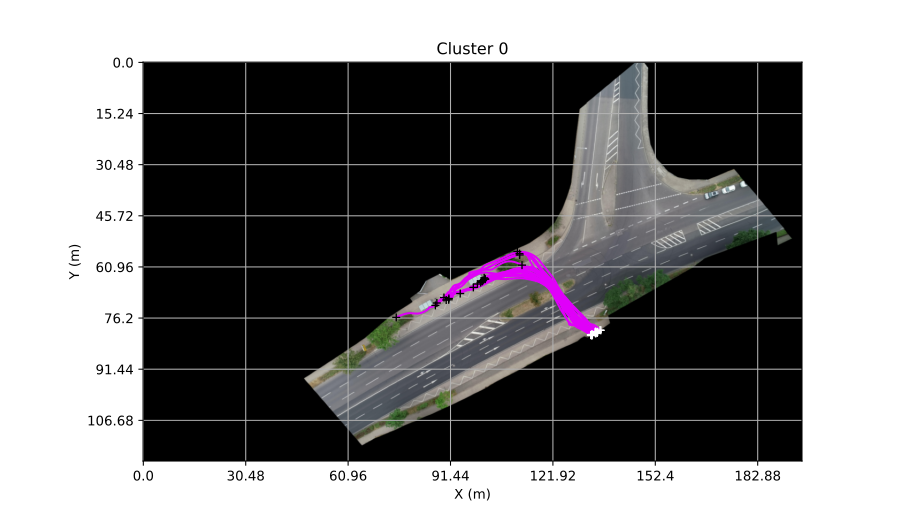} 
		}
		\caption{A2MS method cluster 0}
		\label{fig:4c}
	\end{subfigure}
	\caption{Differences between methods for scenario 0}
	\label{fig:4}
\end{figure}

Figure \ref{fig:4} shows an example of the effect of clustering with initial and final points to then merge the most similar trajectories later can have. In Figure \ref{fig:4a} the result of the pure agglomerative cluster separated cluster 2 can be seen, where it mix an outlier (the white cross in the middle of the trajectory). The A2MS method (A1MS resulted in the same result) not only was able to separate the outliers (figure \ref{fig:4c}) but also merged another pertinent cluster with it.

\begin{figure}[h]
	\centering
	\begin{subfigure}[b]{0.32\linewidth}
		\centering
		\adjustbox{scale=1.25, trim=15mm 15mm 40mm 5mm, clip}{
			\includegraphics[width=2.5\textwidth]{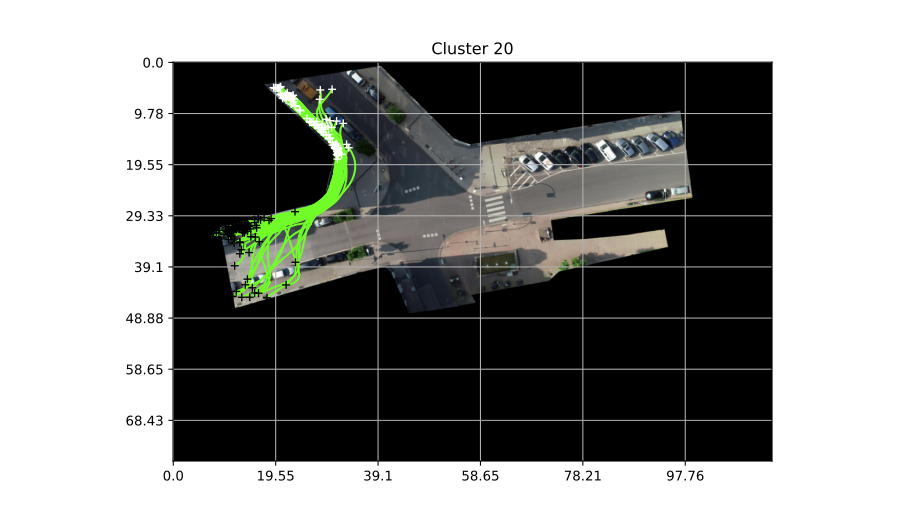} 
		}
		\caption{Agglo. cluster 20}
		\label{fig:5a}
	\end{subfigure}
	\begin{subfigure}[b]{0.32\linewidth}
		\centering
		\adjustbox{scale=1.25, trim=15mm 15mm 40mm 5mm, clip}{
			\includegraphics[width=2.5\textwidth]{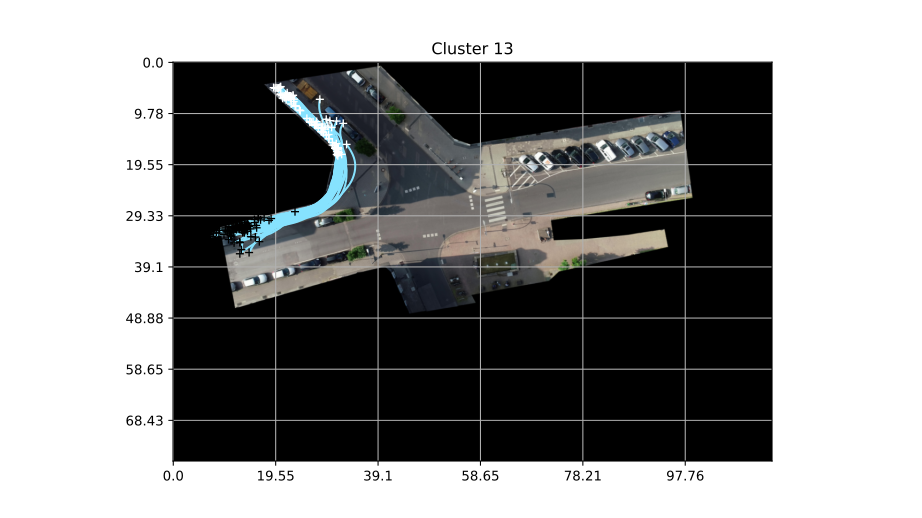} 
		}
		\caption{A2MS. cluster 13}
		\label{fig:5b}
	\end{subfigure}
%	\begin{subfigure}[b]{0.32\linewidth}
%		\centering
%		\adjustbox{scale=1.25, trim=15mm 15mm 40mm 5mm, clip}{
%			\includegraphics[width=2.5\textwidth]{img/a2ms_cl14.png} 
%		}
%		\caption{A2MS cluster 14}
%		\label{fig:5c}
%	\end{subfigure}
	\caption{Differences between methods for scenario 2}
	\label{fig:5}
\end{figure}

Differences between pure agglomerative against both mean-shift post-processing one are clearly visible, but in scenario 2 the results are more similar. This is probably due to the high number of samples, with helped the pure agglomerative method to sift through the outliers, but as it can be seen in Figure \ref{fig:5}, not enough; it can also be seen the efficacy of the A2MS method to remove mixed clusters.

Concerning the comparison with the PAM and dissimilarity methods, one can see that they are inefficient in both fronts being evaluated here: take more time to calculate and do not produce tight clusters, specially because of the outliers present in the scene. In \cite{demoura2023extraction}, these methods were used to discover the different maneuvers of vehicles, but it must be highlighted that car's behaviors are much more constrained than pedestrians (and thus prone to outliers) and that the few outliers observed were removed before execution. 

\subsection{Cyclists}
\label{subsec:4.3}

The cyclist data was acquired at the same intersections than the pedestrians. From the set of trajectories given by the dataset the approximate cyclist behavior can be considered as somewhat between cars and pedestrians, with a constrained movement, but still able to access multiple parts of the road environment. For scenario 1 results (Table \ref{tab:4}) the main distinction that can be made from the pure agglo. and its modification is the opposite of what was observed with pedestrians.

\begin{table}[h]
	\caption{Results for inD dataset scenario 1} 
	\label{tab:4}
	\centering
	\scriptsize
	\adjustbox{max width=\columnwidth}{
		\begin{tabular}{c|cccccc}
			\toprule
			\textbf{Method} & \textbf{Time} & \textbf{DB}	& \textbf{Slh.} & \textbf{Spr.} & \textbf{Best $n_k$} & \textbf{Nb trajs.}\\
			\midrule
			Agglo. 	& 0.0036 s	& 0.6452			& 0.7746			& 0.5748 			& 28 (12) 	& 96.19\% (16)\\
			A2MS 	& 5.44 s	& \textbf{0.4818}	& \textbf{0.8139}	& \textbf{0.4854}	& 26 (12)	& 94.52\% (23)\\
			A1MS 	& 3.86 s	& 0.5199			& 0.8090			& 0.5365			& 26 (12) 	& 95.95\% (17)\\
			PAM 	& 44.50 s	& 0.6040			& 0.0794			& 0.7684			& 38 (35)	& 99.29\% (3)\\
			Dissi. 	& 60.80	s	& 0.6666			& 0.1560			& 0.8571			& 42 (34)	& 98.10\% (8)\\
			\bottomrule
		\end{tabular}
	} 
\end{table}

\begin{figure}[h!]
	\centering
	\begin{subfigure}[b]{0.33\columnwidth}
		\centering
		\adjustbox{scale=1.25, trim=15mm 12.5mm 25mm 5mm, clip}{
			\includegraphics[width=2\textwidth]{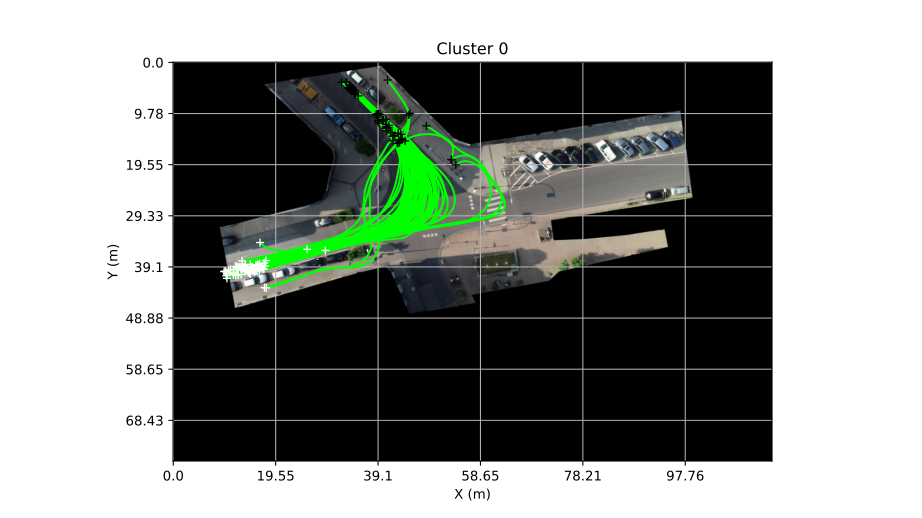} 
		}
		\caption{Agglo. cluster 0}
		\label{fig:8a}
	\end{subfigure}
	\begin{subfigure}[b]{0.32\columnwidth}
		\centering
		\adjustbox{scale=1.25, trim=15mm 12.5mm 25mm 5mm, clip}{
			\includegraphics[width=2\textwidth]{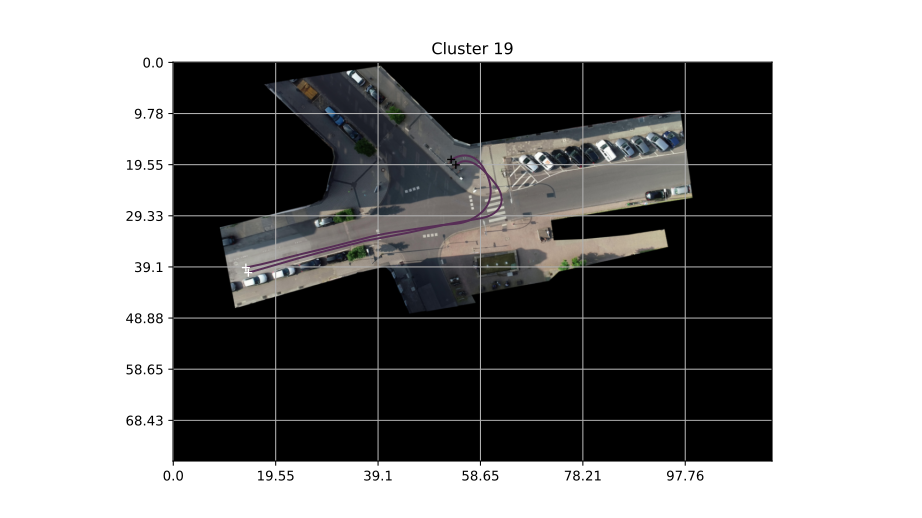} 
		}
		\caption{A2MS cluster 19}
		\label{fig:8b}
	\end{subfigure}
%	\begin{subfigure}[b]{0.32\columnwidth}
%		\centering
%		\adjustbox{scale=1.25, trim=15mm 12.5mm 25mm 5mm, clip}{
%			\includegraphics[width=2\textwidth]{img/a2ms_cl32_cyc.png} 
%		}
%		\caption{A2MS cluster 32}
%		\label{fig:8c}
%	\end{subfigure}
	\caption{Clustering differences for scenario 2}
	\label{fig:8}
\end{figure}

Some of the trajectories were split between multiple clusters for the pure agglomerative method while for the A*MS (meaning both A1MS and A2MS) methods these clusters could be merged together, specially in scenario 2. There are other instances of this same behavior in different clusters as well. Both A*MS methods can attribute their superior scores to the ability to remove outliers from clusters, as illustrated in Figure \ref{fig:8}.

\begin{table}[h]
	\caption{Results for inD dataset scenario 2} 
	\label{tab:5}
	\centering
	\scriptsize
	\adjustbox{max width=\columnwidth}{
		\begin{tabular}{c|cccccc}
			\toprule
			\textbf{Method} & \textbf{Time} & \textbf{DB}	& \textbf{Slh.} & \textbf{Spr.} & \textbf{Best $n_k$} & \textbf{Nb trajs.}\\
			\midrule
			Agglo. 	& 0.035 s	& 0.7789	& \textbf{0.7286}	& 1.6400 &	54 (40) & 99.16\% (14)\\
			A2MS 	& 18.50 s	& \textbf{0.5761}	& 0.6582	& \textbf{1.0211}	& 70 (37)	& 95.21\% (80)\\
			A1MS 	& 11.46 s	& 0.6484			& 0.6316	& 1.3435			& 61 (48) 	& 97.84\% (36)\\
			PAM 	& 1121.17 s	& 0.7331			& 0.2814	& 1.2032			& 60 (59)	& 99.94\% (1)\\
			Dissi. 	& 983.17 s	& 0.8621			& 0.3115	& 1.3177			& 59 (56)	& 99.82\% (3)\\
			\bottomrule
		\end{tabular}
	} 
\end{table}

\subsection{Vehicles}
\label{subsec:4.4}

\begin{figure}[h]
	\centering
%	\begin{subfigure}[b]{0.49\columnwidth}
%	\centering
%	\adjustbox{scale=0.8, trim=30mm 7mm 15mm 18mm, clip}{
%		\includegraphics[width=2\textwidth]{img/a_cl0.png} 
%	}
%	\caption{Agglo. method clust. 0}
%	\label{fig:9a}
%	\end{subfigure}
%	\hfill
%	\begin{subfigure}[b]{0.49\columnwidth}
%		\centering
%		\adjustbox{scale=0.8, trim=30mm 7mm 15mm 18mm, clip}{
%			\includegraphics[width=2\textwidth]{img/a2ms_cl2.png} 
%		}
%		\caption{A2MS method cluster 2}
%		\label{fig:9b}
%	\end{subfigure}

	\begin{subfigure}[b]{0.49\columnwidth}
		\centering
		\adjustbox{scale=0.8, trim=30mm 7mm 15mm 18mm, clip}{
			\includegraphics[width=2\textwidth]{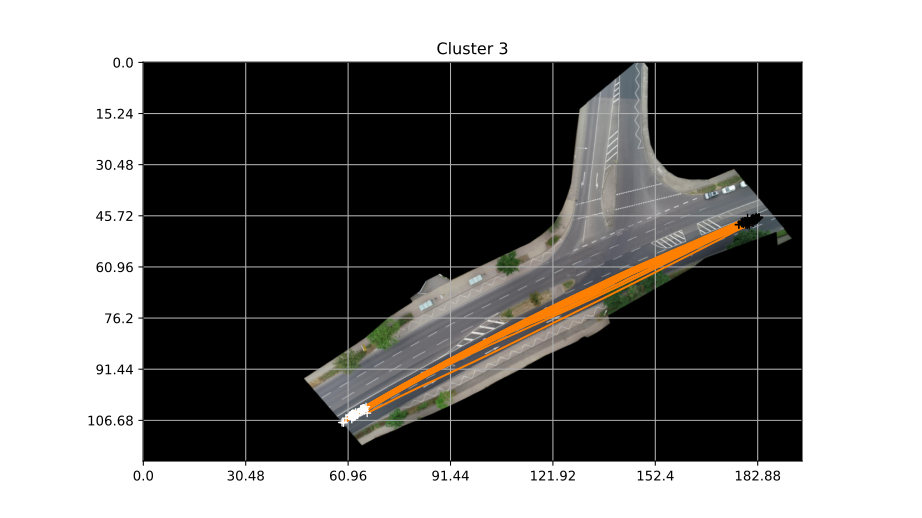} 
		}
		\caption{A2MS method cluster 3}
		\label{fig:9c}
	\end{subfigure}
	\begin{subfigure}[b]{0.49\columnwidth}
		\centering
		\adjustbox{scale=0.8, trim=30mm 7mm 15mm 18mm, clip}{
			\includegraphics[width=2\textwidth]{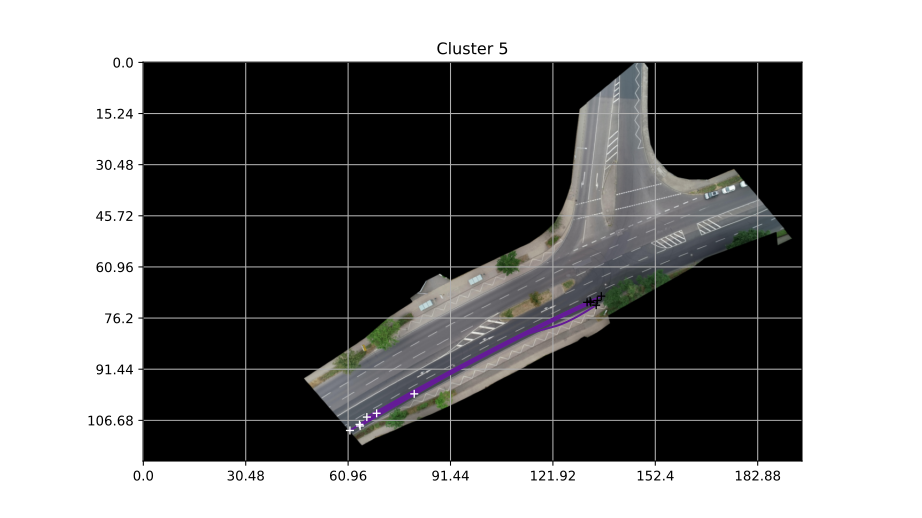} 
		}
		\caption{A2MS method cluster 5}
		\label{fig:9d}
	\end{subfigure}
	\caption{Problems with the A2MS merge when removing outliers}
	\label{fig:9}
\end{figure}

For vehicles the volume of data increases, with the addition of inD dataset scenario 3 and the entire rounD dataset. Since the movements for vehicles are very constrained there are almost none eccentric behavior, hence the goal here is to eliminate all the erroneous samples; for example, trajectories that end at the middle of the intersection. In some cases this was possible, notably on scenarios 1 and 2 for the inD dataset, however, in scenario 0 one maneuver got separated as the result of the mean-shift and merge mechanism for the A2MS method (Figure \ref{fig:9}). Beyond that all other maneuvers for 0 were correctly determined.

\begin{table}[h]
	\caption{Results for inD dataset scenario 0} 
	\label{tab:6}
	\centering
	\scriptsize
	\adjustbox{max width=\columnwidth}{
		\begin{tabular}{c|cccccc}
			\toprule
			\textbf{Method} & \textbf{Time} & \textbf{DB}	& \textbf{Slh.} & \textbf{Spr.} & \textbf{Best $n_k$} & \textbf{Nb trajs.}\\
			\midrule
			Agglo. 		& 0.042 s	& 1.6403			& \textbf{0.9014}	& 0.4538 			&	15 (9) 		& 99.67\% (6)	\\
			A2MS 		& 6.22 s	& \textbf{0.9366}	& 0.8615		 	& \textbf{0.1850}	&	23 (10)		& 99.23\% (14)	\\
			A1MS	 	& 3.67 s	& 1.1694			& 0.6796			& 0.3095			&	22 (13) 	& 99.45\% (10)	\\
			PAM 		& 50.07 s 	& 2.5422			& 0.6283			& 0.6289			& 	13 (13)		& 100\%	(0)		\\
			Dissi. 		& 51.69 s	& 1.9416			& 0.3239			& 0.4938			&	15 (12)		& 99.84 (3)\%	\\
			\bottomrule
		\end{tabular}
	} 
\end{table}

For scenario 1 the agglo., A2MS and A1MS were spot on, with the sole difference that two clusters detected by the agglo and the A1MS are actually outliers and were rejected by A2MS before they formed clusters.

\begin{table}[h]
	\caption{Results for inD dataset scenario 1} 
	\label{tab:7}
	\centering
	\scriptsize
	\adjustbox{max width=\columnwidth}{
		\begin{tabular}{c|cccccc}
			\toprule
			\textbf{Method} & \textbf{Time} & \textbf{DB}	& \textbf{Slh.} & \textbf{Spr.} & \textbf{Best $n_k$} & \textbf{Nb trajs.}\\
			\midrule
			Agglo. 		& 0.10 s	& 1.3758			& 0.8942			& 0.5533 			& 19 (14) 	& 99.79\% (5)\\
			A2MS	 	& 7.74 s	& \textbf{0.8631}	& \textbf{0.9044}	& \textbf{0.2221}	& 20 (12)	& 99.23\% (28)\\
			A1MS	 	& 4.60 s	& 1.2074			& 0.8260			& 0.4431			& 18 (14) 	& 99.19\% (29)\\
			PAM 		& 134.08 s	& 1.5372			& 0.7375			& 0.4658			& 19 	& 100.0\%	\\
			Dissi. 		& 149.07 s	& 2.2906			& 0.7046			& 0.5292			& 20	& 100.0\%	\\
			\bottomrule
		\end{tabular}
	} 
\end{table}

\begin{figure}[!h]
	\centering
	
	\begin{subfigure}[b]{0.49\columnwidth}
		\centering
		\adjustbox{scale=1, trim=37mm 11mm 25mm 11mm, clip}{
			\includegraphics[width=2\textwidth]{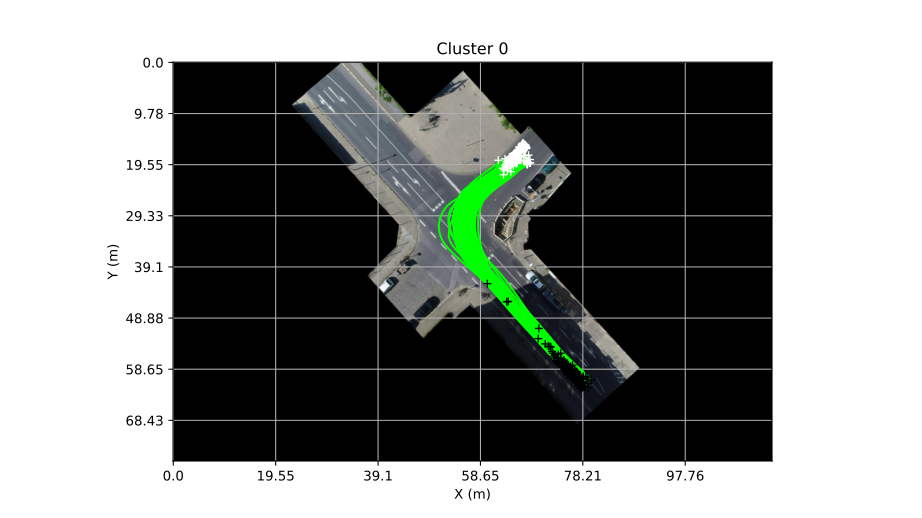} 
		}
		\caption{A2MS method cluster 0}
		\label{fig:11c}
	\end{subfigure}
	\begin{subfigure}[b]{0.49\columnwidth}
		\centering
		\adjustbox{scale=1, trim=25mm 25mm 35mm 6mm, clip}{
			\includegraphics[width=2\textwidth]{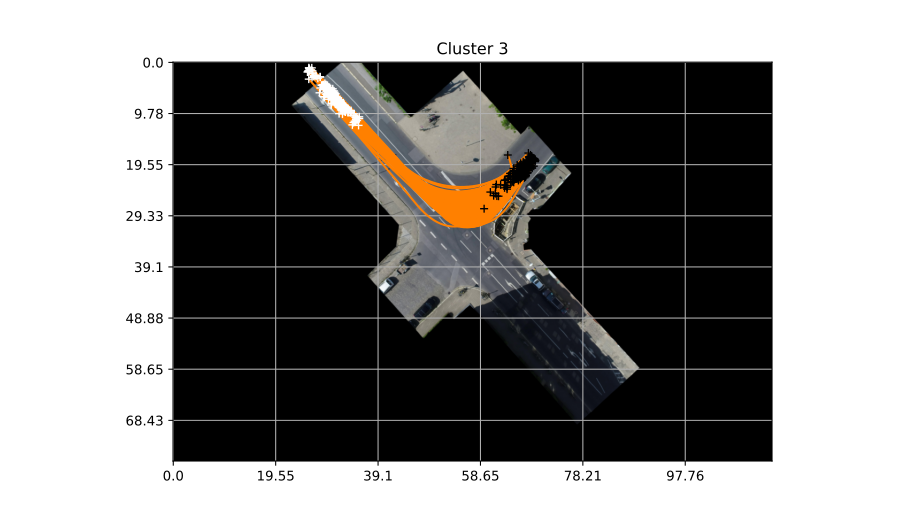} 
		}
		\caption{A2MS method cluster 3}
		\label{fig:11d}
	\end{subfigure}

	\begin{subfigure}[b]{0.32\columnwidth}
		\centering
		\adjustbox{scale=1, trim=27mm 7mm 20mm 8mm, clip}{
			\includegraphics[width=2\textwidth]{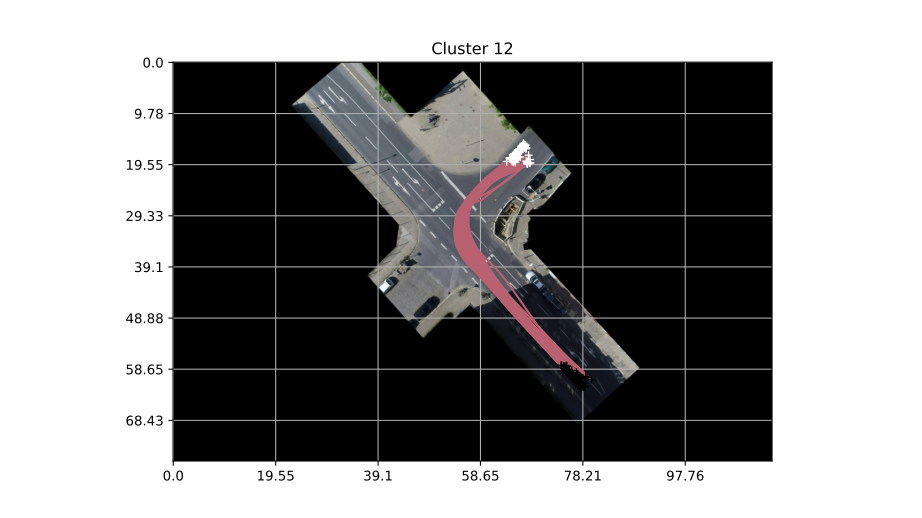} 
		}
		\caption{PAM cluster 12}
		\label{fig:11e}
	\end{subfigure}
	\begin{subfigure}[b]{0.32\columnwidth}
		\centering
		\adjustbox{scale=1, trim=27mm 7mm 20mm 8mm, clip}{
			\includegraphics[width=2\textwidth]{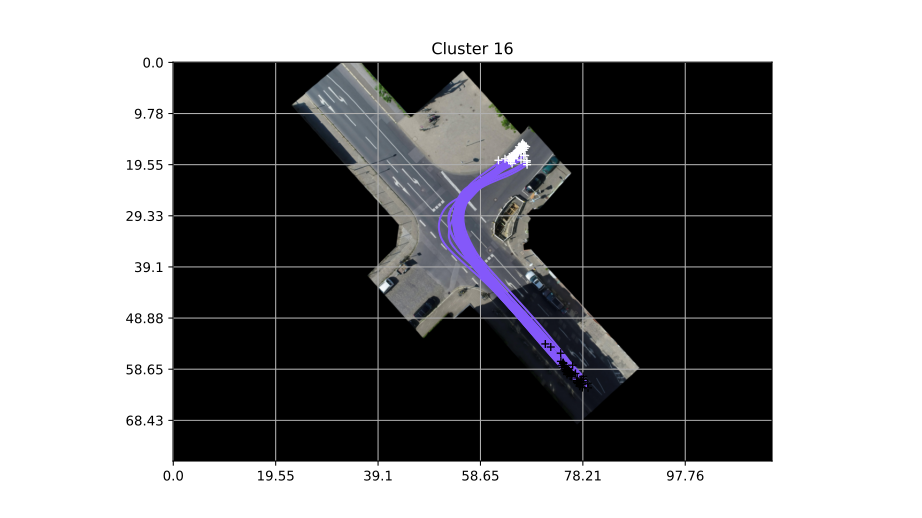} 
		}
		\caption{PAM cluster 16}
		\label{fig:11f}
	\end{subfigure}
	\begin{subfigure}[b]{0.33\columnwidth}
		\centering
		\adjustbox{scale=1, trim=27mm 7mm 20mm 8mm, clip}{
			\includegraphics[width=2\textwidth]{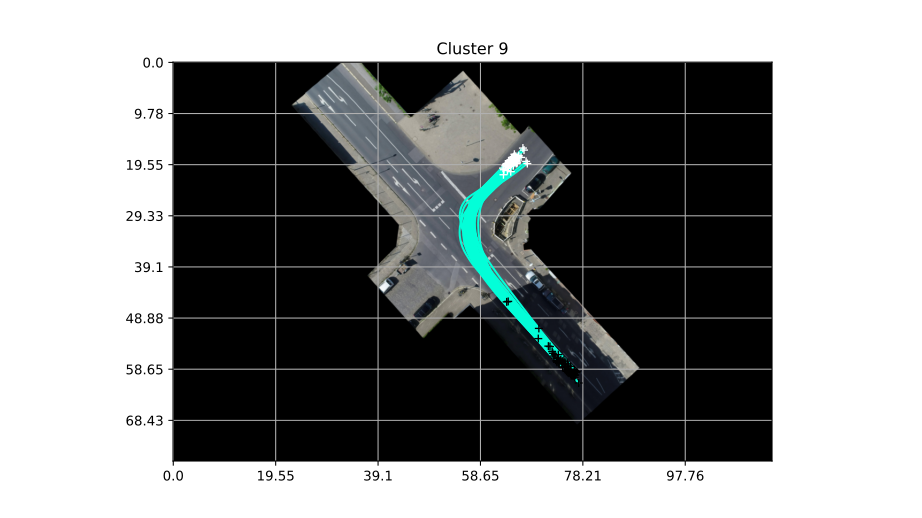} 
		}
		\caption{PAM cluster 9}
		\label{fig:11g}
	\end{subfigure}
	\caption{Outliers (Agglo.) and clusters from A2MS and PAM}
	\label{fig:11}
\end{figure}

As for the PAM and dissi methods, they split different maneuvers into different clusters: Figure \ref{fig:11c} represents the cluster that was divided into Figures \ref{fig:11e}, \ref{fig:11f}, \ref{fig:11g}, which explains huge the disparity shown in Table \ref{tab:7}. 

\begin{figure}[!h]
	\centering
	\begin{subfigure}[b]{0.49\columnwidth}
		\centering
		\adjustbox{scale=0.8, trim=23mm 15mm 40mm 7mm, clip}{
			\includegraphics[width=2\textwidth]{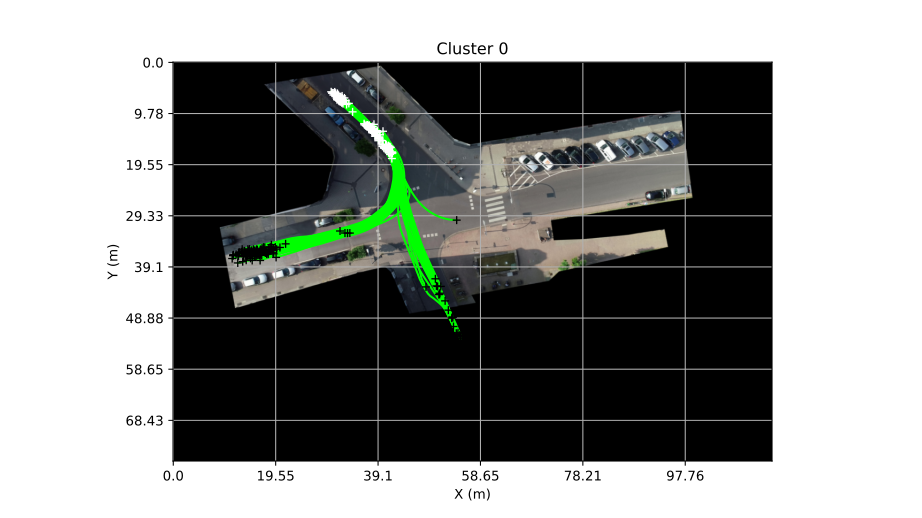} 
		}
		\caption{Agglo. clust. 0}
		\label{fig:12a}
	\end{subfigure}
	\hfill
	\begin{subfigure}[b]{0.49\columnwidth}
		\centering
		\adjustbox{scale=0.8, trim=22mm 15mm 20mm 10mm, clip}{
			\includegraphics[width=2\textwidth]{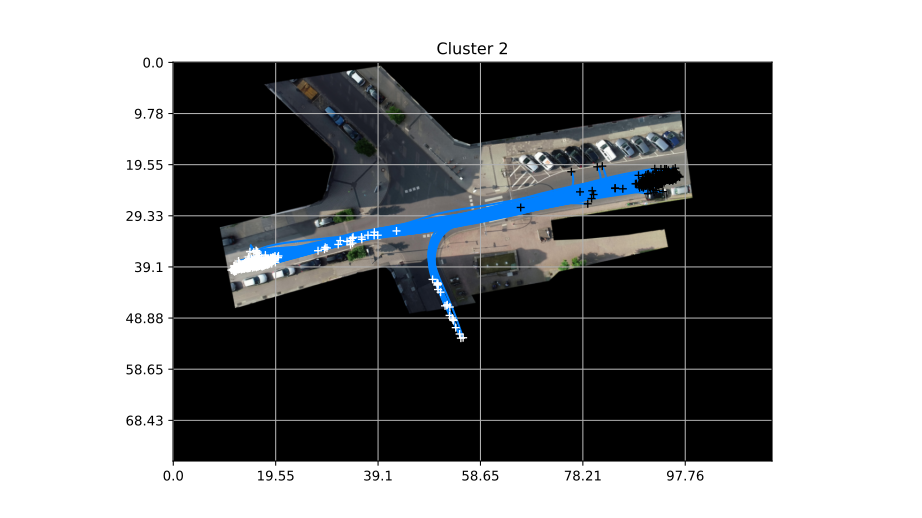} 
		}
		\caption{Agglo. cluster 2}
		\label{fig:12b}
	\end{subfigure}

%	\begin{subfigure}[b]{0.49\columnwidth}
%		\centering
%		\adjustbox{scale=0.8, trim=23mm 20mm 40mm 7mm, clip}{
%			\includegraphics[width=2\textwidth]{img/a2ms_cl5_2.png} 
%		}
%		\caption{A2MS cluster 0}
%		\label{fig:12c}
%	\end{subfigure}
%	\begin{subfigure}[b]{0.49\columnwidth}
%		\centering
%		\adjustbox{scale=0.8, trim=22mm 20mm 20mm 10mm, clip}{
%			\includegraphics[width=2\textwidth]{img/a2ms_cl6_2.png} 
%		}
%		\caption{A2MS cluster 3}
%		\label{fig:12d}
%	\end{subfigure}

	\caption{Differences for scenario 2}
	\label{fig:12}
\end{figure}

Table \ref{tab:8} marks is the first time that the A1MS method had a better DB index and its analogue, due to the splitting of a curve maneuver that contained many samples. This is why the DB score is not used to define the best method, even when the number of cores is the same: there is some situations to split a cluster that should be an unit might be beneficial because of the spread (Equation \eqref{eq:10}) calculation. Besides that, again the pure agglomerative cluster is not capable to split maneuvers that share most of their length (Figures \ref{fig:12a} and \ref{fig:12b})

\begin{table}[h]
	\caption{Results for inD dataset scenario 2} 
	\label{tab:8}
	\centering
	\scriptsize
	\adjustbox{max width=\columnwidth}{
		\begin{tabular}{c|cccccc}
			\toprule
			\textbf{Method} & \textbf{Time} & \textbf{DB}	& \textbf{Slh.} & \textbf{Spr.} & \textbf{Best $n_k$} & \textbf{Nb trajs.}\\
			\midrule
			Agglo. 		& 0.056 s	& 1.7995			& 0.8425			& 1.3679 			& 17 (13) 	& 99.81\% (4)\\
			A2MS 		& 8.02 s	& 1.1375			& \textbf{0.8466}	& \textbf{0.7153}	& 22 (19)	& 98.97\% (22)\\
			A1MS	 	& 5.06 s	& \textbf{1.1286}	& 0.8135			& 0.8857			& 15 (19) 	& 99.02\% (21)\\
			PAM 		& 251.27 s	& 1.7625			& 0.3552			& 1.2958			& 24		& 100\% (0)\\
			Dissi. 		& 209.87 s	& 1.9005			& 0.3444			& 1.5167			& 22 (21)	& 99.95\% (1)\\
			\bottomrule
		\end{tabular}
	} 
\end{table}

For scenario 3 the A1MS actually is the better option, but there is only a difference of two samples classified differently from this case.

\begin{table}[h!]
	\caption{Results for inD dataset scenario 3} 
	\label{tab:9}
	\centering
	\scriptsize
	\adjustbox{max width=\columnwidth}{
		\begin{tabular}{c|cccccc}
			\toprule
			\textbf{Method} & \textbf{Time} & \textbf{DB}	& \textbf{Slh.} & \textbf{Spr.} & \textbf{Best $n_k$} & \textbf{Nb trajs.}\\
			\midrule
			Agglo. 		& 0.014 s	& \textbf{1.9011}	& \textbf{0.9285}	& 0.2949 			& 18 (10) 	& 99.27\% (2)\\
			A2MS	 	& 4.46 s	& 2.3577			& 0.8696		 	& 0.2314			& 20 (9)	& 99.18\% (9)\\
			A1MS	 	& 1.98 s	& 2.1899			& 0.8671			& \textbf{0.2289}	& 19 (9) 	& 99.18\% (9)\\
			PAM 		& 237.22 s 	& 3.1968			& 0.6639			& 0.5730			& 19 (12)	& 99.82\% (2)\\
			Dissi. 		& 64.44 s 	& 3.1112			& 0.6513			& 0.5730			& 18 (12)	& 99.82\% (2)\\
			\bottomrule
		\end{tabular}
	} 
\end{table}

Since the trajectories for the rounD dataset are fairly different in length and direction the DTW distance measure is able to really account trajectories from different maneuvers, as it can be seen in table \ref{tab:10}.

\begin{table}[h]
	\caption{Results for rounD dataset scenario 0} 
	\label{tab:10}
	\centering
	\scriptsize
	\adjustbox{max width=\columnwidth}{
		\begin{tabular}{c|cccccc}
			\toprule
			\textbf{Method} & \textbf{Time} & \textbf{DB}	& \textbf{Slh.} & \textbf{Spr.} & \textbf{Best $n_k$} & \textbf{Nb trajs.}\\
			\midrule
			Agglo. 		& 0.57 s	& 0.4237			& \textbf{0.9470}	& 0.1185		 	& 22 (14) 	& 99.79\% (8)\\
			Agglo. 2MS 	& 10.45 s	& \textbf{0.2817}	& 0.6721		 	& \textbf{0.1135}	& 19 (24)	& 99.71\% (11)\\
			Agglo. MS 	& 5.72 s	& 0.3223			& 0.7633			& 0.1224			& 22 (22) 	& 99.63\% (14)\\
			PAM 		& 329.96 s	& 0.4174			& 0.5618			& 0.5193			& 25 (24)	& 99.97\% (1)\\
			Dissi. 		& 424.83 s 	& 0.6701			& 0.7870			& 0.1184			& 24 (16)   & 99.97\% (1)\\
			\bottomrule
		\end{tabular}
	} 
\end{table}

\begin{figure}[!h]
	\centering
	\begin{subfigure}[b]{0.32\columnwidth}
		\centering
		\adjustbox{scale=1.25, trim=22mm 5mm 13mm 9mm, clip}{
			\includegraphics[width=2\textwidth]{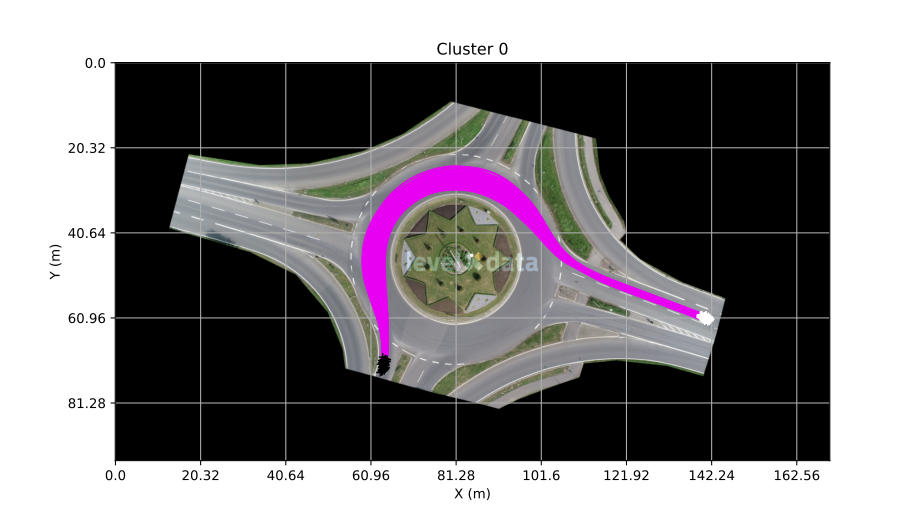} 
		}
		\caption{Agglo. cluster 0}
		\label{fig:13a}
	\end{subfigure}
	\begin{subfigure}[b]{0.32\columnwidth}
		\centering
		\adjustbox{scale=1.25, trim=22mm 5mm 13mm 9mm, clip}{
			\includegraphics[width=2\textwidth]{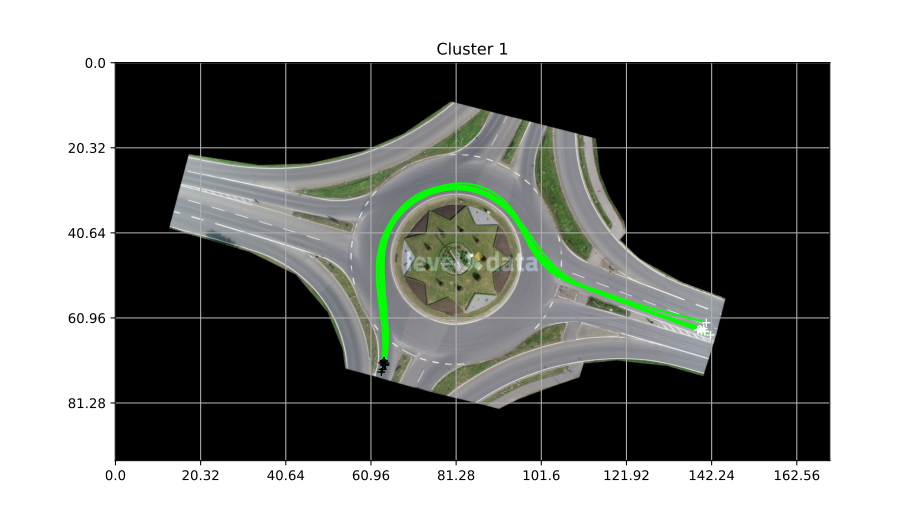} 
		}
		\caption{A2MS cluster 19}
		\label{fig:13b}
	\end{subfigure}
	\begin{subfigure}[b]{0.32\columnwidth}
		\centering
		\adjustbox{scale=1.25, trim=22mm 5mm 13mm 9mm, clip}{
			\includegraphics[width=2\textwidth]{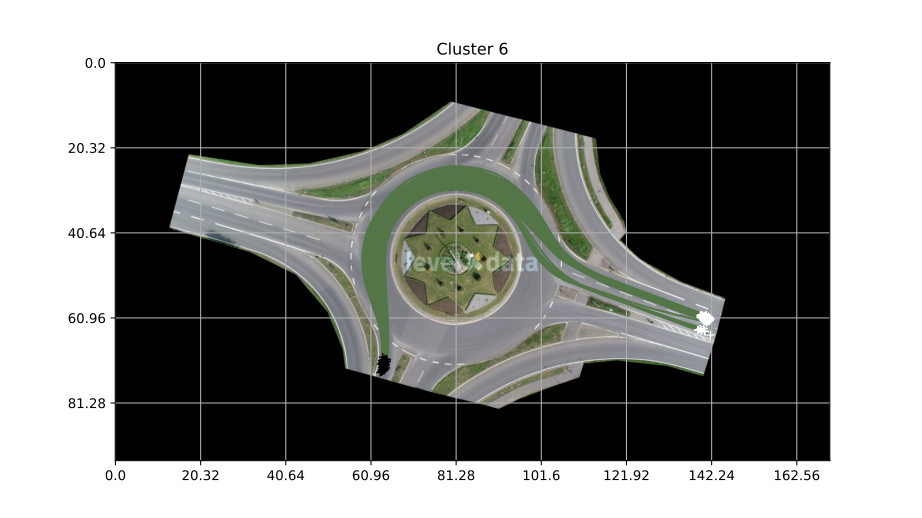} 
		}
		\caption{A2MS cluster 32}
		\label{fig:13c}
	\end{subfigure}
	\caption{Clustering differences for scenario 1}
	\label{fig:13}
\end{figure}

But there is something that are not as salient in the short turns in the intersection from inD dataset but is in this case. Given the size of the roundabout the position in which the vehicles execute the trajectory becomes a discriminating parameter, i.e. the clusters also accounted if they are on the inside or outside (Figures \ref{fig:13a} and \ref{fig:13b} comparing with \ref{fig:13c}) together with the lane in which the vehicle ends or starts its trajectory. This difference mostly impacted the merge step given that now there is a lateral distance through the curve that is bigger than the spread over both clusters. If this division is appropriate or not is in the eye of the beholder.

As a general comment, for the current use-case, the method A2MS proved vastly better than any other, which was made clear by the spread on cluster score defined here. It captures the tightness of each cluster much better than the DB index, which ends up being translated as clusters with few to none outliers in their midst. 

\section{Conclusion}

A new method to cluster trajectories, A2MS, together with a metric defined for the trajectory clustering case, the spread on cluster, were proposed and tested with the datasets inD and rounD. Using an hierarchical clustering combined with DTW distance measure and a as cluster distribution measure, A2MS proved to be the most efficient in the tested methods to produce tight and concentrated clusters with minimal number of outliers. 

The immediate next step is to use this clustering method in conjunction with the longitudinal approach proposed by \cite{demoura2023extraction} to extract drivers' behaviors from real data. But more broadly the method proposed here has multiple uses, from prepare data to learning tasks for planning, decision-making or prediction to even the study of traffic flow in a predetermined zone. Ultimately, this method allows in the future to collect data to train a representation of trajectories so that comparisons could be made with trajectories from different road configurations.

\addtolength{\textheight}{-12cm}   % This command serves to balance the column lengths

\bibliographystyle{IEEEtran}
\bibliography{iv_refs}

\end{document}